\documentclass[cameraready]{Interspeech}

\title{Beyond Word Error Rate: Auditing the Diversity Tax in Speech Recognition through Dataset Cartography}
\author[affiliation={1}, orcid=0009-0008-9180-5117]{Ting-Hui}{Cheng}
\author[affiliation={2}, orcid=0000-0001-5527-5798]{Line H. }{Clemmensen}
\author[affiliation={1}, orcid=0000-0002-4017-1280]{Sneha}{Das}

\address{
    $^1$ Department of Applied Mathematics and Computer Science, Technical University of Denmark, Denmark \\
    $^2$ Department of Mathematical Sciences, University of Copenhagen, Denmark 
}

\email{tiche@dtu.dk, lkhc@math.ku.dk, sned@dtu.dk}


\usepackage{comment}
\usepackage{multirow}
\usepackage{adjustbox}
\usepackage{makecell}

\begin{document}

\maketitle

\begin{abstract}
    Automatic speech recognition~(ASR) systems are predominantly evaluated using the Word Error Rate~(WER). However, raw token-level metrics fail to capture semantic fidelity and routinely obscures the `diversity tax', the disproportionate burden on marginalized and atypical speaker due to systematic recognition failures. In this paper, we explore the limitations of relying solely on lexical counts by systematically evaluating a broader class of non-linear and semantic metrics. To enable rigorous model auditing, we introduce the sample difficulty index~(SDI), a novel metric that quantifies how intrinsic demographic and acoustic factors drive model failure. By mapping SDI on data cartography, we demonstrate that metrics EmbER and SemDist expose hidden systemic biases and inter-model disagreements that WER ignores. Finally, our findings are the first steps towards a robust audit framework for {\it prospective} safety analysis, empowering developers to audit and mitigate ASR disparities {\it prior to deployment}.\footnote{The evaluation framework and analysis code will be made publicly available after decisions.}
\end{abstract}

\section{Introduction}
Word Error Rate (WER) serves as the primary and most commonly used performance metric for evaluating automatic speech recognition~(ASR) systems. WER computes the normalized edit distance between the predicted and reference transcripts and is commonly used for benchmarking ASR systems. In our informal survey of Interspeech papers published between 2023 and 2025 containing the keyword `ASR' and reporting performance results (Table~\ref{tab:exe}), we identified 305 papers. Among them, 86.6\% used WER as an evaluation metric, while fewer than 40\% considered other metrics~\cite{kim2021evaluating}. Furthermore, 180 papers relied exclusively on WER, and only 84 papers employed multiple evaluation metrics. WER is predominantly used, often as the sole metric to benchmark ASR models and gauge their readiness to deployment. This heavy reliance on a single metric raises critical questions regarding its adequacy across diverse acoustic, linguistic and demographic contexts. 

Prior work has highlighted several limitations of WER~\cite{zheng2026towards}, particularly its imperfect alignment with human judgment~\cite{thennal2025advocating}. These limitations of WER are also evident in samples from our study, when different types of errors yield identical WER scores, as shown in Table~\ref{tab:exe}. 
Table~\ref{tab:exe} lists the other commonly used metrics, their usage percentage and examples of how they compare to WER. 
Among these, the next widely used metric is Character Error Rate (CER)~\cite{kurimo06_interspeech}. More recently, alternative evaluation measures have been proposed to better capture different aspects of the recognition quality~\cite{kim2021evaluating, roux2022qualitative}. 
These approaches aim to incorporate linguistic or semantic information into evaluation. For example, WIL typically produces higher values in cases of word substitution. SemDist captures semantic differences and EmbER, which incorporates semantic similarity, fluctuates based on the contextual relevance of the errors rather than simple edit distance. 
These observations highlight that relying solely on WER may therefore provide an incomplete and skewed assessment of ASR performance and should be supplemented by additional evaluation measures~\cite{patel2025evaluate, phukon2025aligning}, and further reinforce the need to move towards multi-dimensional evaluation framework, tied to the application and context. 

\begin{table}[!t]
\centering
\scriptsize
\setlength{\tabcolsep}{2pt} 
\begin{tabular}{llrrrrrr}

\toprule
\multirow{4}{*}{\textbf{Example}}
& \textbf{Metric} & \textbf{WER} & \textbf{CER} & \textbf{MER} & \textbf{WIL} & \textbf{EmbER} & \textbf{SemDist} \\
\cmidrule(lr){2-8}  
& Level & word & character & word & word & word & sentence \\
& Range & $[0,\infty)$ & $[0,\infty)$ & $[0,100]$ & $[0,100]$ & $[0,\infty)$ & $[0,\infty]$ \\
& Usage\% & 86.56\% & 35.41\% & 0.66\% & 0.33\% & 0.33\% & 1.31\% \\
\midrule

\multicolumn{2}{l}{
\begin{tabular}{@{}r@{\ }l@{}} Ref: & ..go meet.. \end{tabular}} &
 \multirow{2}{*}{\textit{0.014}} & \multirow{2}{*}{\textbf{0.009}} & \multirow{2}{*}{\textit{0.014}} & \multirow{2}{*}{\textit{0.014}} & \multirow{2}{*}{\textit{1.429}} & \multirow{2}{*}{0.073} 
 \\

\multicolumn{2}{l}{\begin{tabular}{@{}r@{\ }l@{}} Pred: & ..go \textcolor{red}{to} meet.. \end{tabular}} &
& & & & & 

 \\

\midrule

\multicolumn{2}{l}{\begin{tabular}{@{}r@{\ }l@{}} Ref: & ..Ask her.. \end{tabular}} &
\multirow{2}{*}{\textit{0.014}} & \multirow{2}{*}{\textit{0.006}} & \multirow{2}{*}{\textit{0.014}} & \multirow{2}{*}{\textit{0.014}} & \multirow{2}{*}{\textit{1.429}} & \multirow{2}{*}{0.188} 

\\

\multicolumn{2}{l}{\begin{tabular}{@{}r@{\ }l@{}} Pred: & ..\textcolor{red}{I} ask her.. \end{tabular}} &
 & & & & & 

\\

\midrule

\multicolumn{2}{l}{\begin{tabular}{@{}r@{\ }l@{}} Ref: & ..\textbf{a} snack for.. \end{tabular}} &
\multirow{2}{*}{\textit{0.014}} & \multirow{2}{*}{\textit{0.006}} & \multirow{2}{*}{\textit{0.014}} & \multirow{2}{*}{\textit{0.014}} & \multirow{2}{*}{\textbf{\textit{1.449}}} & \multirow{2}{*}{0.113} 
 \\

\multicolumn{2}{l}{\begin{tabular}{@{}r@{\ }l@{}}Pred : & ..snack for.. \end{tabular}} &
 & & & & & 

 \\

\midrule

\multicolumn{2}{l}{\begin{tabular}{@{}r@{\ }l@{}} Ref: & ..\textbf{the} store.. \end{tabular}} &
\multirow{2}{*}{\textit{0.014}} & \multirow{2}{*}{\textit{0.006}} & \multirow{2}{*}{\textit{0.014}} & \multirow{2}{*}{\textbf{\textit{0.029}}} & \multirow{2}{*}{0.145} & \multirow{2}{*}{0.084} 

 \\

\multicolumn{2}{l}{\begin{tabular}{@{}r@{\ }l@{}} Pred : & ..\textcolor{red}{this} store.. \end{tabular}}&
&  & & & & 

 \\

\midrule

\multicolumn{2}{l}{\begin{tabular}{@{}r@{\ }l@{}} Ref: & ..plastic \textbf{snake}.. \end{tabular}} &
\multirow{2}{*}{\textit{0.014}} & \multirow{2}{*}{\textit{0.006}} & \multirow{2}{*}{\textit{0.014}} & \multirow{2}{*}{\textbf{\textit{0.029}}} & \multirow{2}{*}{\textbf{\textit{1.449}}} & \multirow{2}{*}{\textbf{0.395}} 

 \\

\multicolumn{2}{l}{\begin{tabular}{@{}r@{\ }l@{}} Pred: & ..plastic \textcolor{red}{snack}.. \end{tabular}} &
& & & & & 

 \\

\bottomrule
\end{tabular}
\caption{Overview of ASR evaluation metrics, including unit level, value range, and usage in Interspeech papers over the past three years, alongside examples of reference and predicted transcriptions from the Speech Accent Archive. Bold indicates the worst (highest) score in each column. Italics highlight values that are identical to at least one other value in the same column, showing where metrics fail to distinguish between corpora.}\label{tab:exe}
\vspace{-1cm}
\end{table}

However, there is still a lack of systematic investigation into how these metrics relate and interact with one another. An additional challenge in evaluating ASR performance is understanding how dataset characteristics influence metric behavior. While prior research has shown that speaker traits and content characteristics are encoded and differentiated within ASR representations~\cite{11003448}, variations in speaker distribution, linguistic complexity, or acoustic conditions may alter error patterns and consequently affect how different metrics assess system performance. Examining whether evaluation measures behave consistently across datasets with differing properties is therefore an important yet underexplored question.

We move ASR evaluation beyond aggregate scores to audit item-level model failures. Our core contributions are: (1) exposing the redundancy and complementarity of standard ASR metrics; (2) quantifying metric elasticity across diverse dataset characteristics; and (3) introducing the Sample Difficulty Index (SDI) to map intrinsic acoustic and demographic traits directly to extrinsic model failure, revealing how metric sensitivity fluctuates across marginalized or atypical speakers.

\section{Experiment Setup}

In this work, we evaluate four common ASR models on 5 datasets, over diverse acoustic and demographic characteristics. 

\begin{figure}[htpb]
    \centering
    \includegraphics[width=0.85\linewidth]{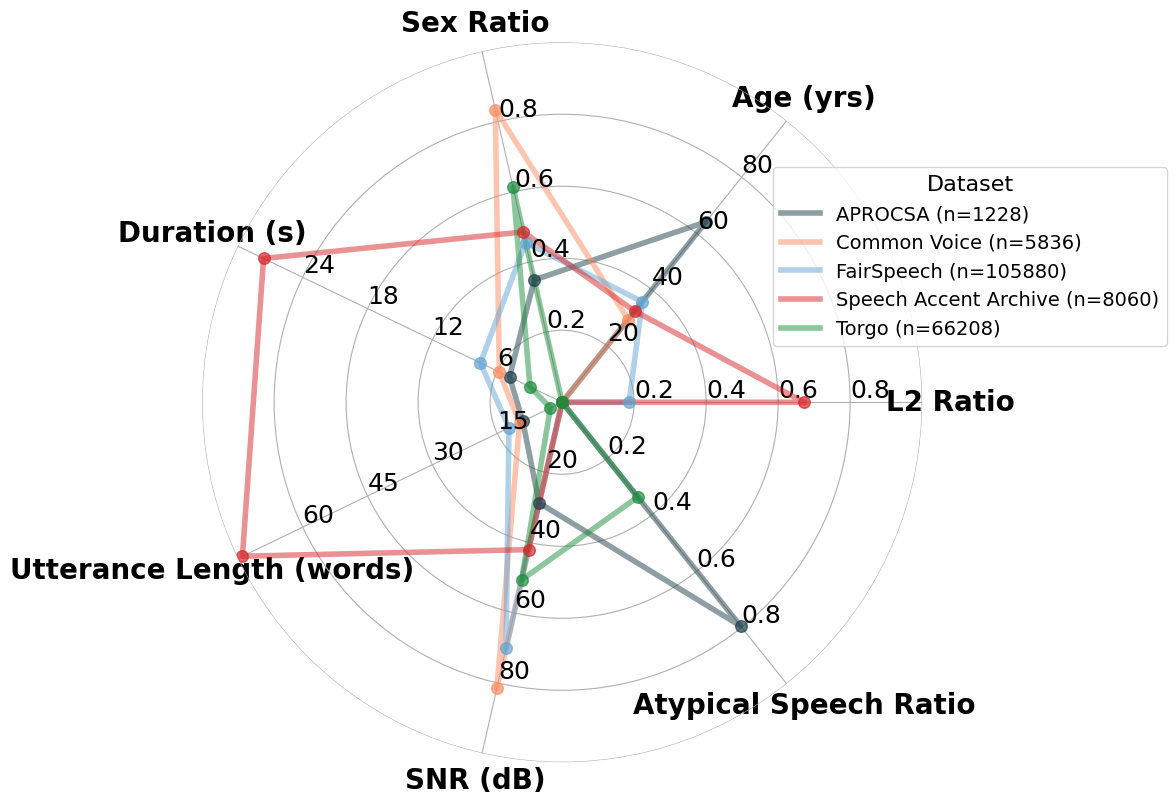}
    \caption{Characteristics of datasets as averages or ratios.}
    \label{fig:chars}
    \vspace{-.5cm}
\end{figure}

\begin{itemize}
  \item Models: Wav2Vec2-Base-960h~\cite{baevski2020wav2vec}, Whisper-Small~\cite{radford2023robust}, STT En Fast Conformer-CTC Large~\cite{rekesh2023fast}, MMS-1b-all~\cite{pratap2024scaling}
  \item Datasets: TORGO~\cite{rudzicz2012torgo}, Speech Accent Archive
  ~\cite{weinberger2015speech}, APROCSA~\cite{ezzes2022open}, Common Voice~\cite{ardila2020common}, Fair-Speech dataset~\cite{veliche2024towards}
  \item ASR Evaluation metrics: Word Error Rate (WER), Character Error Rate (CER)~\cite{kurimo06_interspeech}, Match Error Rate (MER)~\cite{morris2004and}, Word Information Lost (WIL)~\cite{morris2002information}, Embedding Error Rate (EmbER)~\cite{roux2022qualitative}, and Semantic Distance (SemDist)~\cite{kim2021evaluating}.
\end{itemize}
Figure \ref{fig:chars} illustrates the characteristic profiles derived from the average of all samples within each dataset. The ratio is defined as the proportion of specific groups in the dataset, including male speakers, L2 speakers, and those with atypical speech. 
Because this study aims to perform an explanatory audit rather than train a predictive model, we utilize the full corpus ($N=185\times10^3$) in our investigation.

\begin{figure*}[!tbh]
    \centering
    \includegraphics[width=.9\textwidth]{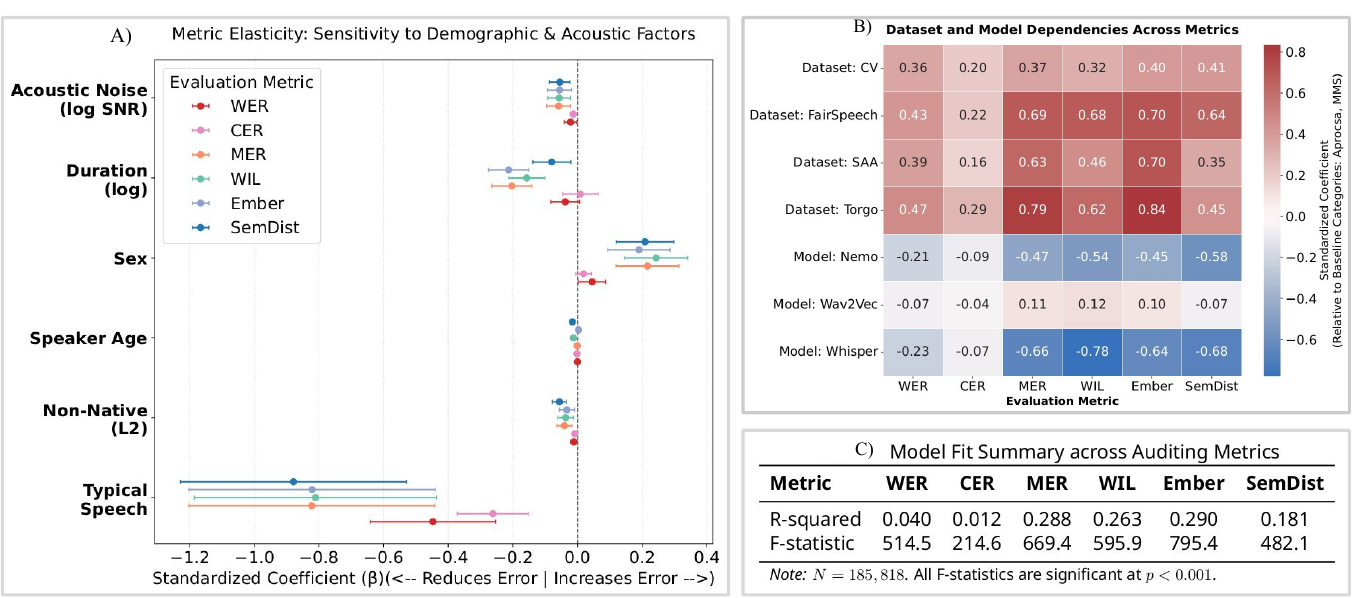}
    \caption{A) Mean and std error of $\beta, \alpha$ coefficients from the fixed effect~(FE) model for the demographic and acoustic characteristics. B) Mean Coefficients from FE of models and datasets; C) Summary of the FE model fit statistics across the six metrics.}
    \label{fig:FE}
    \vspace{-.5cm}
\end{figure*}

\section{Methodology}

\subsection{Metric Complementarity Analysis}
To investigate the complementarity of the 6 evaluation metrics, we apply Principal Component Analysis (PCA) to examine their underlying covariance structure. We aggregate evaluation results across all four ASR models and five datasets. All metric values are standardized by removing the mean and scaling to unit variance. 
By examining the factor loadings of the first three principal components, which account for 93.6\% of the variance, we determine whether each metric reflects shared variance with others or captures distinct dimensions of ASR performance.

\subsection{Metric elasticity}
Current state of ASR evaluation is predominantly based on macro-averaging. Mathematically, this assumes that evaluation metrics are primarily a function of the chosen architecture and the corpus it is tested on, effectively reducing the evaluation paradigm to:
$Y^{metric} \sim A(D), $
where $Y^{metric}$ is the performance metric of the model, $A$ is the architecture, and $D$ is the dataset. By reducing datasets to monolithic entities, this approach treats intra-dataset speaker and demographic variance (such as acoustic noise, speaker age, or L1 status) as zero-mean random noise hence masking the diversity tax~\cite{hollands2022evaluating, koenecke2020racial}. 

 Our hypothesis in this paper is that this approach is incomplete and provides an overestimate of the real-world performance and robustness of ASR models. In this paper, we define {\it metric-elasticity as the isolated sensitivity of an ASR metric to specific acoustic or demographic characteristics}. Hence, the Metric Elasticity Audit Framework (MEAF) upgrades evaluation from the static, two dimensional leaderboard into a multi-dimensional audit:
 \vspace{-.15cm}
\begin{equation} 
Y^{metric} \sim A + D + C_{Ac} +  C_{De}, \end{equation}
$C_{Ac}$ and $C_{De}$ being the acoustic and demographic characteristics, together referred to as the dataset characteristics and further defined below. \\
\noindent
\textbf{Dataset characteristics: } Speech datasets exhibit inherent variability in their provided metadata. To systematically quantify the intrinsic properties of each dataset, we use the the following granular dimensions towards a multi-dimensional framework to characterize datasets. 
\begin{itemize}
\item \textbf{SNR (dB)}: Signal-to-Noise Ratio of speech signals estimated using WADA-SNR\cite{kim2008robust} to quantify the audio quality~($x_{\text{snr}}$).

\item \textbf{Sample Duration ($x_{\text{len}}$, log-sec)}: Temporal length of individual segments measures in seconds and transformed to logarithm to address the heavy skew of the distribution.
\item \textbf{Age ($x_{\text{age}}$)}: Age metadata extracted from each dataset. Categorical age bins were mapped to numeric midpoints. Missing values were mean-imputed and flagged with a binary indicator ($x_{\text{miss}}$). Available ages were standardized into Z-scores.
\item \textbf{Demographic variables}: Binary categorical variables representing sex~($sex$), non-native~($L1$) and typical vs. atypical speech~($Typ$). 






\end{itemize}

\noindent
\textbf{Statistical model:} We use speaker-clustered fixed effects regression to isolate and quantify the marginal impact of intersecting demographic (eg: L1-L2 status, atypical speech, sex, age) and acoustic factors. By introducing architecture and dataset as control fixed effects~(FE), the statistical model absorbs systemic baseline variance (eg: disparities in model parameter size), quantifying the pure performance penalty attributable to the speakers themselves. 
\vspace{-.15cm}
\begin{equation}
\begin{split}
Y_{s,i,m}^{\text{metric}} &= \beta_0 + \underbrace{\beta_{\text{snr}} x_{\text{snr}, i} + \beta_{\text{len}} x_{\text{len}, i} + \beta_{\text{age}} x_{\text{age}, i} + \beta_{\text{miss}} x_{\text{Miss}, i}}_{\text{Continuous Slopes}} \\
&\quad + \underbrace{\alpha_{\text{sex(i)}} + \alpha_{L1(i)} + \alpha_{\text{Typ}(i)}}_{\text{Demographic Fixed Effects}} + \underbrace{\gamma_{d(i)} + \delta_{m}}_{\text{Systemic Fixed Effects}} + \epsilon_{s,i,m}, 
\end{split}
\end{equation}
where $\gamma_{d}, \delta_{m}$ are the coefficients of the dataset and model effects, $\alpha$ models the FE intercept shifts for the demographic variables and $\epsilon_{s,i,m}$ is the speaker clustered error term. 
All continuous independent and dependent variables were standardized before modeling, to enable convergence of the statistical models and a comparison of the coefficients over all the 6 metrics.

\subsection{Sample difficulty index (SDI) \& cartography validation}
Using the elasticity weights ($\beta$ \& $\alpha$) derived from the statistical model, we construct the SDI, a metadata-driven scalar that quantifies the compounding impact of the demography and acoustic traits of an utterance. 
\begin{equation}
\text{SDI}_i = \boldsymbol{\beta}^\top \mathbf{x}_i + \sum_{j \in \{\text{sex}, L1, \text{Typ}\}} \alpha_{j(i)},
\end{equation}
$\boldsymbol{\beta}^\top \mathbf{x}_i$ is the dot product of the coefficients and the standardized continuous features (SNR, duration, age) for utterance $i$, and $\alpha_{j(i)}$ are the FE intercepts for the categorical demographic groups.

To extrinsically validate the SDI independent of the regression itself, we project it onto a multi-model cartography map~\cite{swayamdipta}. While conventional dataset cartography~\cite{swayamdipta} plots the training dynamics of a single model across epochs, we adapt this framework to map cross-architecture evaluation dynamics by calculating the mean error and variance across an ensemble of distinct ASR models. Specifically, cartography plots the mean error~($\mu_i^{\text{metric}} =\frac{1}{4}\sum_{m=1}^4 y_{i, m}^{\text{metric}}$) against inter-model disagreement ($\sigma = \sqrt{\frac{1}{4}\sum_{m=1}^4 (y_{i,m}^{\text{metric}}-\mu_i^{\text{metric}})^2}$) and maps regions of sample difficulty based on these two parameters. Because the Cartography coordinates are derived strictly from empirical model behavior, while the SDI is derived strictly from the sample's acoustic and demographic metadata, a strong spatial correlation between the two serves as objective validation. 


\section{Results}

\begin{figure}[h]
    \centering
    \includegraphics[width=0.8\linewidth]{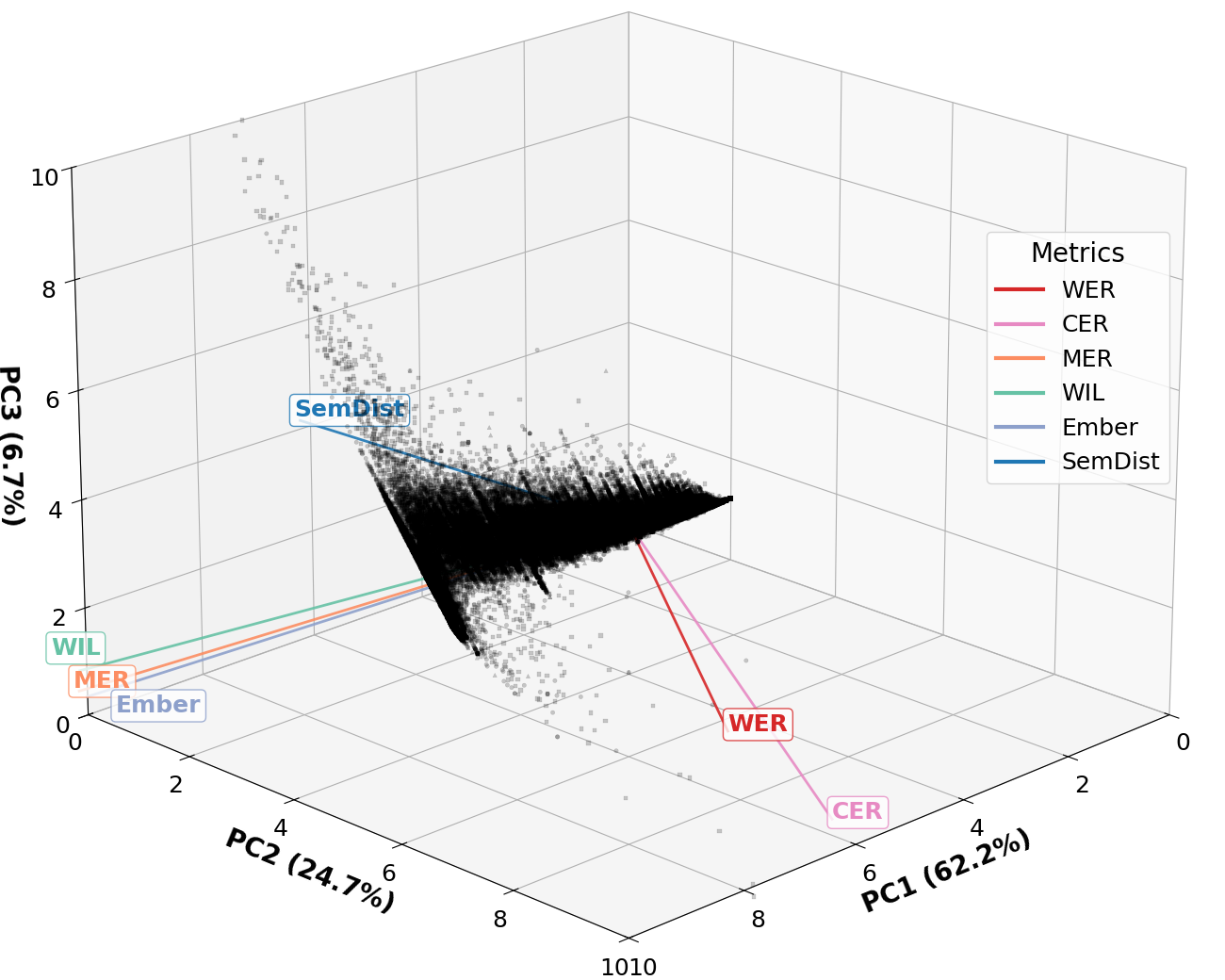}
    \caption{Latent space mapping of ASR performance using Principal Component Analysis. Axes are truncated to highlight the primary variance clusters.}
    \label{fig:PCA}
    \vspace{-.5cm}
\end{figure}

\textbf{Three-way metric divergence:} Figure~\ref{fig:PCA} presents the PCA projection of performance metrics for 3 principal components, accounting for 93.6\% of the variance. Three distinct variable groupings are visible. First, WER and CER follow similar trajectories, though CER diverges from WER along PC2 and PC3. 
 Second, WIL, MER, and EmbER cluster closely
 , which suggests redundancy among these token-level metrics. Finally, SemDist occupies a distinct direction, capturing variance along components not aligned with the other metrics. This separation highlights that SemDist encodes complementary information relative to the other error measures.\\
\noindent
\begin{figure*}[!tbh]
    \centering
    \includegraphics[width=.9\textwidth]{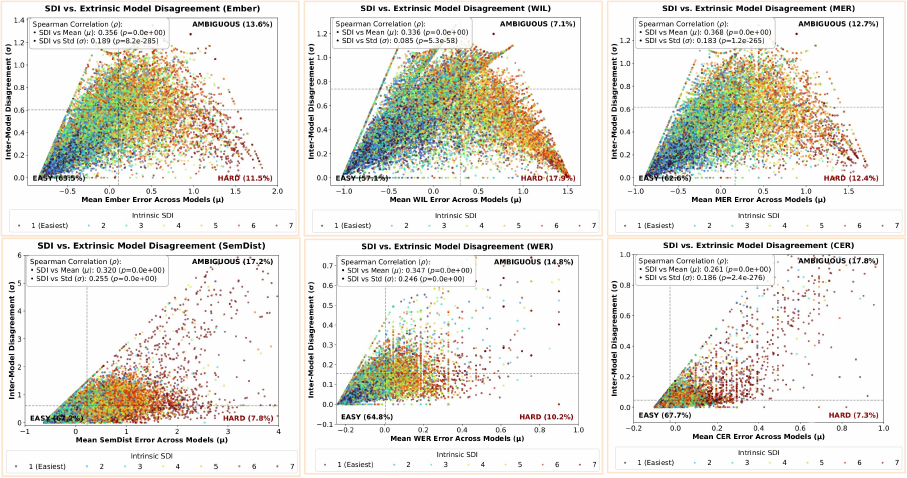}
    \caption{
    Cartography plots mapping mean error ($\mu$) against inter-model disagreement ($\sigma$), colored by SDI decile (1 = Easiest, 10 = Hardest). SDI Deciles divide the dataset's speech samples into ten equal tiers based on their calculated intrinsic difficulty, ranging from 1 (the easiest samples for models to transcribe) to 10 (the hardest).}
    \label{fig:SDIVCartography}
    \vspace{-.5cm}
\end{figure*}
\noindent\textbf{Dataset characteristics influence metrics differently: }Evaluation metrics follow a clear hierarchy of elasticity; while lexical counts (WER, CER) remain relatively stable, non-linear and semantic measures capture significantly more demographic friction, providing a more transparent audit of the `Diversity Tax'. This term refers to the disproportionate cognitive and practical burden placed on users with marginalized or atypical speech characteristics, who must constantly adapt their pronunciation or repeatedly correct transcription errors just to achieve the same baseline utility as majority-demographic users.

Figure~\ref{fig:FE} shows that the evaluation metrics exhibit varying degrees of sensitivity to speaker characteristics. Overall, WER and CER are less sensitive to demographic and acoustic factors, as evidenced by their lower standardized coefficients and and $R^2$ values (0.040 and 0.012, respectively). This suggests that raw lexical error counts are dominated by stochastic noise or unobserved linguistic variables, rather than a systematic coupling to the speaker's profile. 
In contrast, MER, WIL, EmbER, and SemDist exhibit greater elasticity, capturing significant performance fluctuations that reveal a deeper dependency on diverse speaker characteristics. These metrics utilize non-linear normalizations, such as the union of reference and hypothesis in the denominator, making them mathematically more sensitive to the hallucinations and omissions common in atypical or L2 speech. Notably, EmbER shows the highest coupling to metadata, with an $R^2$ of 0.290 and an F-statistic (795.4) nearly four times higher than character, level metrics, confirming its role as a high-sensitivity indicator for demographic friction.

\begin{figure}[!h]
    \centering
    \includegraphics[width=\columnwidth]{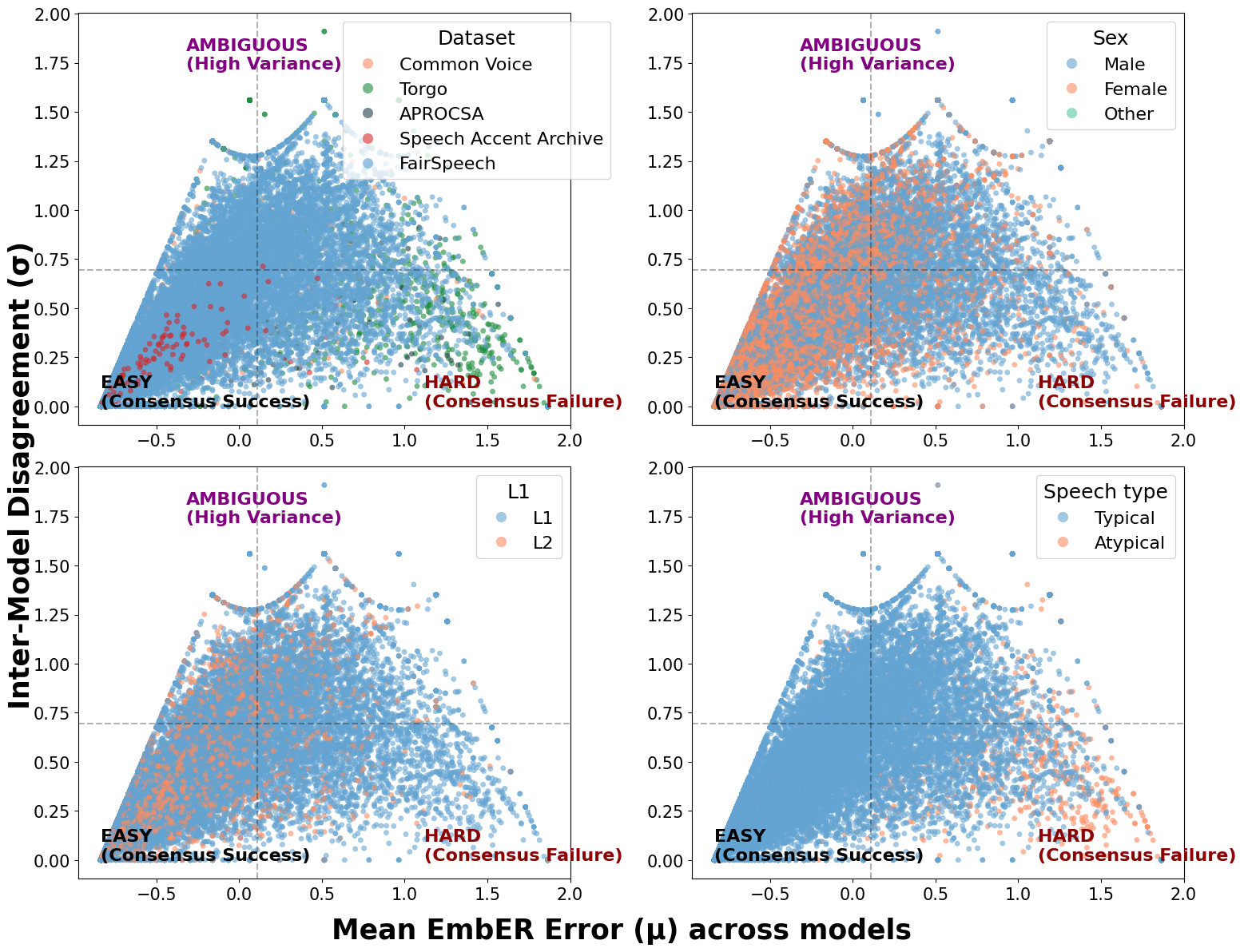}
    \caption{Cartography plots of mean EmbER error ($\mu$) against inter-model disagreement ($\sigma$) across ASR systems using the EmbER metric. The $\mu$ reflects the overall recognition difficulty of each sample, while $\sigma$ captures the extent of ambiguity, indicating how consistently different ASR models perform on the same utterance.}
    \label{fig:cartography}
    \vspace{-.25cm}
\end{figure}

A similar pattern is observed across models and datasets. WER and CER show relatively low dependency on architectural differences, whereas the remaining metrics are more sensitive to such variations, reflecting stronger responsiveness to changes in modeling approaches and data conditions. \\
\noindent
\textbf{Difficulty of samples vs. attributes: }
\label{sec:difficulty}
To examine how different attributes relate to ASR difficulty, we present EmbER cartography plots stratified by demographic and speech characteristics (Figure~\ref{fig:cartography}). Atypical speech samples cluster in regions of high mean error and relatively low inter-model disagreement, indicating that these utterances are challenging for ASR systems. In contrast, samples from female and L2 speakers are concentrated in regions with lower mean error and reduced disagreement, suggesting that these specific female and L2 samples are comparatively easier to transcribe. 

\noindent
\textbf{Validating SDI via dataset cartography: }
Building on the attribute-level difficulty patterns observed in previous section, we next examine whether the proposed Intrinsic SDI captures the empirical difficulty of a sample. Figure~\ref{fig:SDIVCartography} presents
, visually and statistically, higher SDI values consistently and significantly correlate with increased mean error across all metrics. For SemDist, WER, and CER, higher SDI deciles are distinctly associated with greater inter-model disagreement ($\sigma$), pushing these samples into the highly variable `Ambiguous' quadrant. This indicates that for these metrics, intrinsic difficulty yields highly unstable predictions across different models. 

In contrast, metrics like EmbER, MER, and WIL exhibit a strictly linear spatial gradient. Here, low-SDI samples are tightly concentrated in the low-error `Easy' quadrant, while high-SDI samples reliably occupy regions of elevated $\mu$ and $\sigma$. Across all metrics, samples with intermediate SDI deciles successfully map to transitional regions of the cartography space, confirming that the intrinsic SDI serves as a robust proxy for extrinsic model dynamics.

\section{Conclusion}

In this work, we identify three groups of ASR evaluation metrics and show that SemDist, EmbER capture more nuanced transcription failures, providing complementary information to token-level measures. We introduce SDI, a quantitative measure of how intrinsic demographic and acoustic factors drive model performance. By mapping SDI onto dataset cartography, we establish a direct link between specific speaker characteristics and high inter-model disagreement, effectively visualizing the diversity tax in action. Our findings reveal systematic vulnerabilities in ASR systems, offering an audit framework for prospective safety analysis to expose and mitigate performance disparities prior to real-world deployment. \textbf{Limitations:} 1) calculating the SDI relies on explicit metadata \& unobserved linguistic or environmental variables that contribute to inter-model variance may remain unaccounted for. 2) Semantic metrics need future validation for typologically diverse languages.

\section{Generative AI Use Disclosure}
Generative AI tools were used for language editing and stylistic refinement.

\bibliographystyle{IEEEtran}
\bibliography{ref_1}

\end{document}